\documentclass{sig-alternate-2013}

\usepackage{lmodern}

\usepackage[T1]{fontenc}

\usepackage{textcomp}
\usepackage{hyperref}

\usepackage{color}
\usepackage{multirow}

\usepackage{lineno,xcolor}

\newcommand{\figref}[1]{Figure~\ref{fig:#1}}
\newcommand{\tabref}[1]{Table~\ref{tab:#1}}

\newcommand{\secref}[1]{Section~\ref{sec:#1}}

\newcommand{\comment}[1]{{}}

\begin{document}
%

\title{DeepSentiBank: Visual Sentiment Concept Classification with Deep Convolutional Neural Networks }
\numberofauthors{1}
\author{\alignauthor
	Tao Chen$^1$, Damian Borth$^2$, Trevor Darrell$^2$ and Shih-Fu Chang$^1$~ \\ ~ \\
	\affaddr{~~~~$^1$Columbia University, USA~~~~~~~~~~~~~~~~$^2$University of California, Berkeley, CA} \\
    \affaddr{$^1$\{taochen,sfchang\}@ee.columbia.edu~~~~$^2$\{borth@icsi,trevor@eecs\}.berkeley.edu
    }
}
%
%
%
%
%

%

\maketitle
\begin{abstract}

This paper introduces a visual sentiment concept classification method based on deep convolutional neural networks (CNNs). The visual sentiment concepts are adjective noun pairs (ANPs) automatically discovered from the tags of web photos, and can be utilized as effective statistical cues for detecting emotions depicted in the images. Nearly one million Flickr images tagged with these ANPs are downloaded to train the classifiers of the concepts. We adopt the popular model of deep convolutional neural networks which recently shows great performance improvement on classifying large-scale web-based image dataset such as ImageNet. Our deep CNNs model is trained based on Caffe, a newly developed deep learning framework. To deal with the biased training data which only contains images with strong sentiment and to prevent overfitting, we initialize the model with the model weights trained from ImageNet. Performance evaluation shows the newly trained deep CNNs model SentiBank 2.0 (or called DeepSentiBank) is significantly improved in both annotation accuracy and retrieval performance, compared to its predecessors which mainly use binary SVM classification models.

\end{abstract}


\section{Introduction}

The explosive growth of social media and online visual content has motivated the research on large-scale social multimedia analysis. Among these research efforts, understanding the emotion and sentiment in visual media content has attracted increasing attention in research and practical applications.. Images and videos depicting strong sentiments can strengthen the opinion conveyed in the content and more effectively influence the audience. Understanding sentiment expressed in visual content will greatly benefit social media communication and enable broad applications in education, advertisement and entertainment.

Modeling generic visual concepts (nouns) such as ``sky'' and ``dog'' has been studied extensively in computer vision, but modeling adjectives correlated with visual sentiments like ``amazing'' and ``shy'' remains difficult, if not impossible, due to the big ``affective gap'' between the low-level visual features and the high-level sentiment.
Therefore, Borth et al. \cite{borth2013large} proposed a more tractable approach which models sentiment related visual concepts as a mid-level representation to fill the gap. Those concepts are Adjective Noun Pairs (ANPs), such as ``happy dog'' and ``beautiful sky'', which combine the sentimental strength of adjectives and detectability of nouns. Though these ANP concepts do not directly express emotions or sentiments, they were discovered based on strong co-occurrence relationships with emotion tags of web photos, and thus are useful as effective statistical cues for detecting emotions depicted in the images. In \cite{borth2013large} binary SVM classifiers of the ANPs are trained on the whole images, denoted as SentiBank 1.1. Later Chen et al. \cite{chen2014object} improve these classifiers by considering object-based concept localization and leveraging semantic similarity among the concepts.

The dataset for training the visual sentiment concepts involves thousands of categories consisting of about one million images downloaded from Flickr. Recently, Krizhevsky et al. \cite{NIPS2012_4824} show deep convolutional neural networks (CNNs) is able to achieve great classification performance improvement and efficiency on similar datasets such as ImageNet~\cite{deng2012large}. The model has a much larger learning capacity that can be controlled by varying the network depth and breadth, compared to SVM and other learning methods. Its strong assumptions of stationarity of statistics and locality of pixel dependencies about the nature of images are also mostly correct. CNNs are also easier to train than standard feedforward neural networks with layers of similar size, since they have much fewer connections and parameters, with only slightly degraded theoretic performance. CNNs also have the capability to incorporate model weights learned from more general dataset, which can be applied to our case by transferring the model learned over ImageNet to the specialized dataset like SentiBank..

This work introduces SentiBank 2.0, or called DeepSentiBank, a visual sentiment concepts classification model which is trained under Caffe~\cite{Jia13caffe,Jia2014caffe}, a GPU based deep learning framework. We adopt similar CNNs architecture used in \cite{NIPS2012_4824} while training on the ILSVRC2012~\cite{deng2012large} dataset. We find that initializing the model with the model weights trained from ImageNet provides much better performance that training from visual sentiments dataset alone. Performance evaluation and comparisons with its predecessors show the newly trained  DeepSentiBank significantly improves the annotation accuracy in ANP classification as well as moderately improves the ANP retrieval performance.

\section{Related Work}
\subsection{Modeling Sentiment}
Most work on sentiment analysis so far has been based on textual information
\cite{wilson05contextualpolarity, esuli2006sentiwordnet,thelwall2010sentiment}. Sentiment models have been demonstrated to be useful in various applications including human behavior prediction \cite{esuli2006sentiwordnet},
business \cite{pang2008opinion}, and
political science \cite{tumasjan2010predicting}.

Compared to text-based sentiment analysis, modeling sentiment based on images has been much less studied. The most relevant work is \cite{borth2013large}, which proposed to design a large-scale visual sentiment ontology based on Adjective-Noun Pairs (the sentiment modeling is then based on one-vs-all SVMs). Chen et al. \cite{chen2014object} further improve the model by considering object-based concept localization and leveraging semantic similarity among the concepts.

\subsection{Modeling Visual Concepts}
Concept modeling has been widely studied in multimedia \cite{Naphade06, smith2003multimedia}, and computer vision (often referred as ``attributes'') \cite{Ferrari07}.
The concepts being modeled are mostly
objects \cite{smith2003multimedia},
scenes \cite{patterson2012sun},
or activities \cite{fu2012attribute}.
There is work trying to solve the ``fine grained recognition'' task, where the categories are usually organized in a hierarchical structure.
\cite{deng2013fine, duan2012discovering, deng2009imagenet}.
There is also work trying to model ``non-conventional'' concepts or properties of the images, such as image aesthetic and quality
\cite{joshi2011aesthetics, marchesotti11aesthetic},
memorability \cite{isola11memorable}, interestingness \cite{isola11memorable}, and affection/emotions \cite{machajdik10affective, wang2008survey, jia2012can, machajdik10affective, wang2008survey, yanulevskaya08emotioncategorization}. The models are usually trained by SVM and other layer lacking learning methods.

\subsection{Deep Learning}
Deep convolutional networks have been long studied in computer
vision. Successful results on digit recognition using supervised back-propagation networks have been achieved in early research\cite{lecun1989backpropagation}. More recently,
similar networks are applied on large benchmark datasets consisting of more than one million images, such as ImageNet
\cite{deng2012large},
 with competition-winning results \cite{NIPS2012_4824}.

The learned deep representations can be transferred across tasks. It has been extensively studied in an unsupervised
setting \cite{raina2007self,mesnil2012unsupervised}. However, such models in convolutional networks
have been limited to relatively small datasets such
as CIFAR and MNIST, and only achieved modest success in \cite{le2013building}. Sermanet et al. \cite{sermanet2013pedestrian} propose to use unsupervised pre-training, followed by supervised fine-tuning to solve the problem of insufficient training data. Supervised pre-training approach using a concept-bank paradigm \cite{kennedy2006lscom,torresani2010efficient} is also proven successful in computer vision and multimedia settings. It learns the features on large-scale data in a supervised setting, then transfers
them to different tasks with different labels. Recently, Girshick et al. \cite{girshick2014rich} shows that supervised pre-training on a large dataset, followed by domain-adaptive fine-tuning on smaller dataset is an efficient paradigm for scarce data.

\begin{figure*}
\centering
  \includegraphics[width=\linewidth]{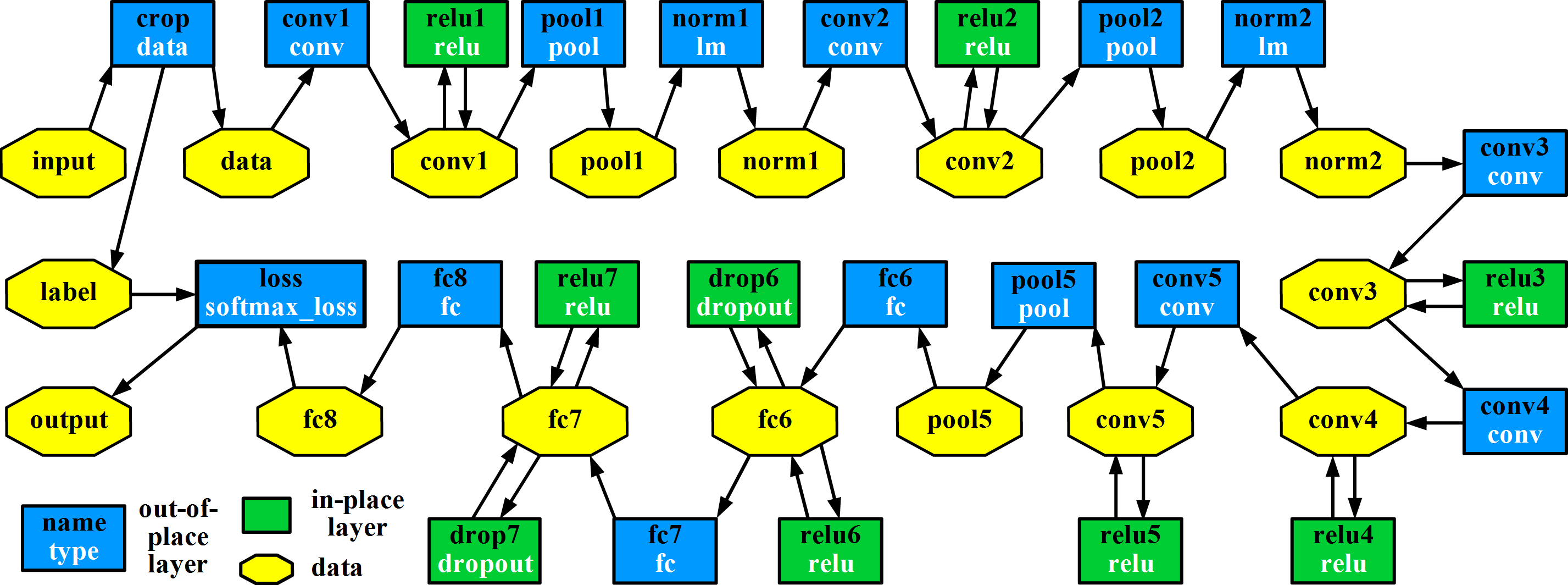}\\
  \caption{The architecture of the deep convolutional neural networks.}\label{fig:pipeline}
\end{figure*}

\section{Visual Sentiment Ontology and Concepts Overview}
In this section, we briefly review the visual sentiment ontology construction in \cite{borth2013large} and define our classification problem.

\subsection{Building Ontology}
The analysis of emotion, affect and sentiment from visual content has become an exciting area in the multimedia community allowing to build new applications for brand monitoring, advertising, and opinion mining. To create an corpora for sentiment analysis on visual content and stimulate innovative research on this challenging issue, a database is constructed by Borth et al. \cite{borth2013large}. This database contains a Visual Sentiment Ontology (VSO) consisting of more than 3,000 adjective noun pairs (ANPs), SentiBank\footnote{Version 1.1 can be downloaded from \\\url{http://visual-sentiment-ontology.appspot.com/}.}, a set of 1,200 trained visual concept detectors providing a mid-level representation of sentiment, and associated training images acquired from Flickr. Construction of the VSO is founded on psychological research by data-driven discovery - for each of the 24 emotions defined in Plutchik's theory \cite{plutchik1980emotion}, images and videos are retrieved from Flickr and YouTube respectively to extract concurrent tags. The set of all adjectives and all nouns is then used to form ANPs such as ``beautiful flowers'' or ``sad eyes''. SentiBank is then trained on the images tagged by these ANPs.

\subsection{Dataset}
\label{sec:dataset}
The database contains a set of Flickr images for training and testing ANP classifiers in SentiBank 1.1. For each ANP, at most 1,000 images tagged with it are downloaded, resulting about one million images for 3,316 ANPs. To train the visual sentiment concept or ANP classifiers, we first filter out the ANPs associated with less than 120 images. 2,089 ANPs with 867,919 images are left after filtering. For each ANP, 20 images are randomly selected for testing, while others are used in training, ensuring at least 100 training images per ANP. To prevent bias in the test set, any training image and test image pair associated with same ANP must not share a same publisher on Flickr. The ANP tags from Flickr users are used as labels for each image. Note those labels may suffer from incompleteness and noisiness, i.e., not all true labels are annotated and sometimes there are falsely assigned labels also. However we do not fix them due to the huge amount of annotation tasks. We use the labels as is and thus will refer to them as pseudo ground truth.

We also build a subset to compare the retrieval performance of different models. This subset only contains images associated with six nouns, namely ``car'', ``dog'', ``dress'', ``face'', ``flower'' and ``food''. These nouns are not only frequently tagged in the social multimedia, but also associated with diverse adjectives to form a large set of ANPs (135 in total). Its training set is the corresponding subset of the full training set. Its test set however, contains 60 manually annotated images for each ANP, where 20 are positive and 40 are negative. The retrieval performance is evaluated by the average precision on the ranking result of the 60 test images for each ANP. For this dataset, we will compare the new DeepSentiBank with an earlier version of SentiBank using object-based localization, called SentiBank 1.5R (indicating region based SentiBank) \cite{chen2014object}.

\section{Deep Convolutional Neural Networks Solution}
\subsection{Introduction of Caffe}

Caffe is a deep learning framework developed by taking full account of cleanliness, readability, and speed. It was created by Jia \cite{Jia13caffe}, and is in active development by the Berkeley Vision and Learning Center (BVLC) and by community contributors. Caffe is released under the BSD 2-Clause license \footnote{\url{http://caffe.berkeleyvision.org/}}.

Using Caffe for deep learning programming has multiple advantages. Its clean architecture enables rapid deployment. Networks are specified in simple config files, with no hard-coded parameters in the code. Switching between CPU and GPU is as simple as setting a flag ¨C so models can be trained on a GPU machine, and then used on commodity clusters.

\subsection{CNN Architecture}

Here we describe the overall architecture of the deep convolutional neural networks for training the visual sentiment concept classification model, SentiBank 2.0 or DeepSentiBank. The architecture mostly follows \cite{NIPS2012_4824}. As depicted in \figref{pipeline}, the net
contains eight main layers (conv or fc) with weights; the first five are convolutional and the other three are fully-
connected. The output of the last fully-connected layer is fed to a 2089-way softmax which produces a distribution over the 2089 class labels. The network maximizes the average across training instances of the log-probability
of the correct label under the prediction distribution by multinomial logistic regression.
The kernels of the second, fourth, and fifth convolutional layers are connected only to half of kernel
maps in the previous layer. The kernels of the third
convolutional layer are connected to all kernel maps in the second layer. The neurons in the fully-connected layers are connected to all neurons in the previous layer. Following \cite{nair2010rectified}, the Rectified Linear Units (ReLUs)
non-linearity  $f(x) = max(0, x)$ is applied to the output of every convolutional and fully-connected layer. Overlapping max-pooling layers (pool) follow the first, second and fifth ReLU layers (relu). The pooling layer consists of a grid of pooling
units spaced 2 pixels apart, each summarizing a neighborhood of size $3 \times 3$ centered at the location
of the pooling unit. Local response normalization layers (lm)
follow the first and second pooling layers. The response-normalized activity $b^i_{
x,y}$ is given by
the expression
$$ b^i_{x,y} = a^i_{x,y}/\left( k+\alpha\sum^{min(N-1,i+n/2)}_{j=max(0,i-n/2)}(a^i_{x,y})^2 \right) ^\beta$$
where $a^i_{x,y}$ is the activity of a neuron computed by max-pooling, the sum runs over $n$ ``adjacent'' kernel maps at the same spatial position, and $N$ is the total
number of kernels in the layer. The constants $k = 2, n = 5, \alpha = 10^{-4}$, and $\beta = 0.75$. The dropout layers (dropout) are applied in the first two fully-connected layers.

\begin{table}
  \centering
  \caption{The input/output data size (left) and the layer shape (right) for each layer.}\label{tab:data}
  \vspace{0.1in}

\begin{tabular}{|c|c|c|c|c|}
\cline{1-2}
name & size
\\
\cline{1-2}
input & $3 \times 256 \times 256$
\\
\cline{1-2}
data & $3 \times 227 \times 227$
\\
\cline{1-2}
conv1 & $96 \times 55 \times 55$
\\
\cline{1-2}\cline{4-5}
pool1 & $96 \times 27 \times 27$ &  & name & shape
\\
\cline{1-2}\cline{4-5}
norm1 & $96 \times 27 \times 27$ && conv1 & $96\times 3\times 11\times 11$
\\
\cline{1-2}\cline{4-5}
conv2 & $256 \times 27 \times 27$ && conv2 & $256\times 48\times 5\times 5$
\\
\cline{1-2}\cline{4-5}
pool2 & $256 \times 13 \times 13$ && conv3 & $384\times 256\times 3\times 3$
\\
\cline{1-2}\cline{4-5}
norm2 & $256 \times 13 \times 13$ && conv4 & $384\times 192\times 3\times 3$
\\
\cline{1-2}\cline{4-5}
conv3 & $384 \times 13 \times 13$ && conv5 & $256\times 192\times 3\times 3$
\\
\cline{1-2}\cline{4-5}
conv4 & $384 \times 13 \times 13$ && fc6 & $ 4096\times 9216$
\\
\cline{1-2}\cline{4-5}
conv5 & $256 \times 13 \times 13$ && fc7 & $4096\times 4096$
\\
\cline{1-2}\cline{4-5}
pool5 & $256 \times 6 \times 6$ && fc8 & $2089\times 4096$
\\
\cline{1-2}\cline{4-5}
fc6 & $4096$
\\
\cline{1-2}
fc7 & $4096 $
\\
\cline{1-2}
fc8 & $2089 $
\\
\cline{1-2}
label & $1$
\\
\cline{1-2}
output & $1$
\\
\cline{1-2}
\end{tabular}
\end{table}

The input/output data size and the layer shape for each layer is shown in \tabref{data}. All training and test images are first normalized to $256 \times 256$ without keeping the aspect ratio. To prevent overfitting, we apply data augmentation consists of generating image translations and horizontal reflections. We do this by extracting random $227\times227$ patches (and their horizontal reflections) from the
$256 \times 256$ images and training our network on these extracted patches. The first convolutional layer filters the $227\times227\times3$ input image with 96 kernels of size $11\times11\times3$
with a stride of 4 pixels. The second convolutional layer takes as input the (pooled and response-normalized) output of the first convolutional layer and filters it with 256 kernels of size $5\times5\times48$.
The third, fourth, and fifth convolutional layers are connected to one another without pooling or normalization. The third convolutional layer has 384 kernels of size $3\times3\times256$ connected to the (normalized and pooled) outputs of the second convolutional layer. The fourth
convolutional layer has 384 kernels of size $3\times3\times192$ , and the fifth convolutional layer has 256
kernels of size $3\times3\times192$. The fully-connected layers have 4096 neurons each.

\subsection{Learning Details}

The regression objective is minimized by stochastic gradient descent
with a batch size of 256 examples, momentum of 0.9, and
weight decay of 0.0005. The small weight decay here is not only a regularizer by also reduces the model's training error.

Due to insufficient data and the bias to images with strong sentiment, training on our dataset may suffer from overfitting. Since our dataset is from the same domain of ImageNet, it is promising to use fine-tuning. We initialized the weights by the model trained from ILSVRC2012 except the top layer. the pre-trained model can be downloaded from \url{http://caffe.berkeleyvision.org/getting_pretrained_models.html}.
The learning rate is initialized at 0.001. Regarding the full forward-backward pass of each batch as an iteration, we run a total of 250,000 iterations (about 77 epochs). We divide learning rate by 10 after every 100,000 iterations (about 20 epochs).

For comparison, we also train a similar model without fine-tuning. We initialize the weights in each layer from a zero-mean Gaussian distribution with standard deviation 0.01. We initialize the neuron biases in the second, fourth, and fifth convolutional layers,
as well as in the fully-connected hidden layers with the constant 0.1, and in the remaining layers with the constant 0. The learning rate is initialized at 0.01.

During testing, we center crop the test images into $227 \times 227$, apply forward propagation with the trained model weights and use the softmax as predicted probability of each concept.

\section{Experimental results}

\subsection{Computation Speed}
Our experiment is done on a single server machine with 16-core dual Intel E5-2650L processor, 64GB memory and a nVidia K20 GPU.
The training over 826,806 images takes about 9 days and testing over 41,780 test images takes about 6 minutes. The maximum memory used is 42GB, and storing data takes 300GB disk space.

\subsection{Performance and Comparisons}
We evaluate the new classification model by both annotation accuracy (measured by the percentage of images that have the pseudo ground truth label in top detected concepts) and retrieval performance (measured by mean average precision).

\subsubsection{Annotation accuracy}

\begin{figure}
\centering
  \includegraphics[width=\linewidth]{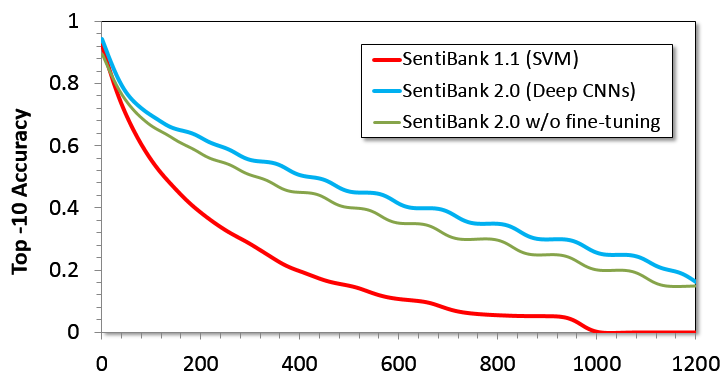}\\
  \caption{The curves of ranked top-10 accuracy per ANP of different approaches. The curves have been smoothed.}\label{fig:accuracy}
\end{figure}

\begin{figure*}
\centering
  \includegraphics[width=\linewidth]{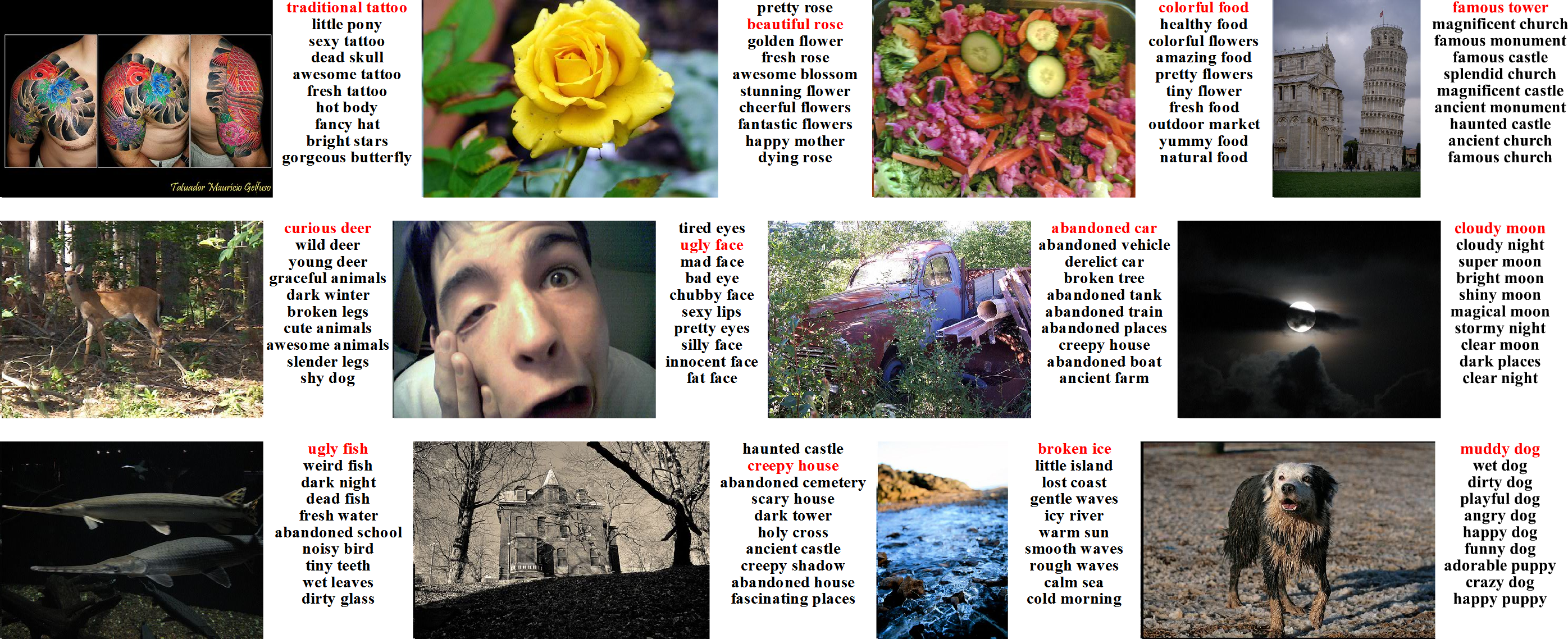}\\
  \caption{Examples of top 10 concepts detected from test images by the fine-tuned DeepSentiBank model. The red concepts are the pseudo ground truth concepts. Credits of images (from top to bottom, from left to right): \textcopyright Mauricio Gelfuso, Bob Wright, fraKara, Melanie Bateman, Photographs-n-Memories, Matt Swanson, Twan Goossens, S Debras, Erin Nichols, Yael Levine, 7 Years Later... and Anda Stavri of Flickr.}\label{fig:top}
\end{figure*}

The annotation accuracy is evaluated on the full test set of 2,089 ANPs mentioned in \secref{dataset} and measured by top-k accuracy - the percentage of images that have the pseudo ground truth label in top k  detected concepts. Top-1, 5, 10 accuracies of each and all ANPs are computed and compared among fine-tuned deep CNNs model (SentiBank 2.0), deep CNNs model without fine-tuning, and SentiBank 1.1 \cite{borth2013large}. The overall accuracies are listed in \tabref{accuracy}. Different from genetic visual concepts, some visual sentiment concepts can be very abstract, such as ``terrible crime'' and ``strong community''. Such ANPs usually have very low classification performance, and are meaningless to be included in the classifiers library for generating mid-level sentiment related features. Thus it is important to compare the performances of ANPs with acceptable detectability. Similar to \cite{borth2013large}, for each approach, we select top 1,200 ANPs ranked by Top-10 accuracy. Note different approach will produce different ANP subsets. The overall accuracies for these subsets are also shown in \tabref{accuracy}. \figref{accuracy} shows the curve of ranked top-10 accuracy per ANP for each subset. According to the table and the figure it is clear that the CNNs-based approaches greatly outperform the SVM based approach, with as much as 370\% performance gain on top-1 accuracy, 200\% on top-5, and 150\% on top-10. Fine-tuned model is also 14\textasciitilde25\% better than the one without fine-tuning. \figref{top} shows some examples of top detected concepts from test images by the fine-tuned model. It shows that despite the serious problem of incomplete and incorrect labels in our dataset, the top detected concepts can still be accurate. Since the pseudo ground truth labels may not be correct, thus the top-5 and top-10 accuracies are more appropriate than top-1 accuracy. We also realize an important reason for the performance boost is that the SVM based SentiBank trains binary classifiers, rather than a general multi-label classification approach. Such binary classification setting is more suitable for retrieval, instead of annotation. Thus, in the next section, we will evaluate the performance of DeepSentiBank in terms of image retrieval.

\subsubsection{Retrieval performance}

\begin{figure}
\centering
  \includegraphics[width=\linewidth]{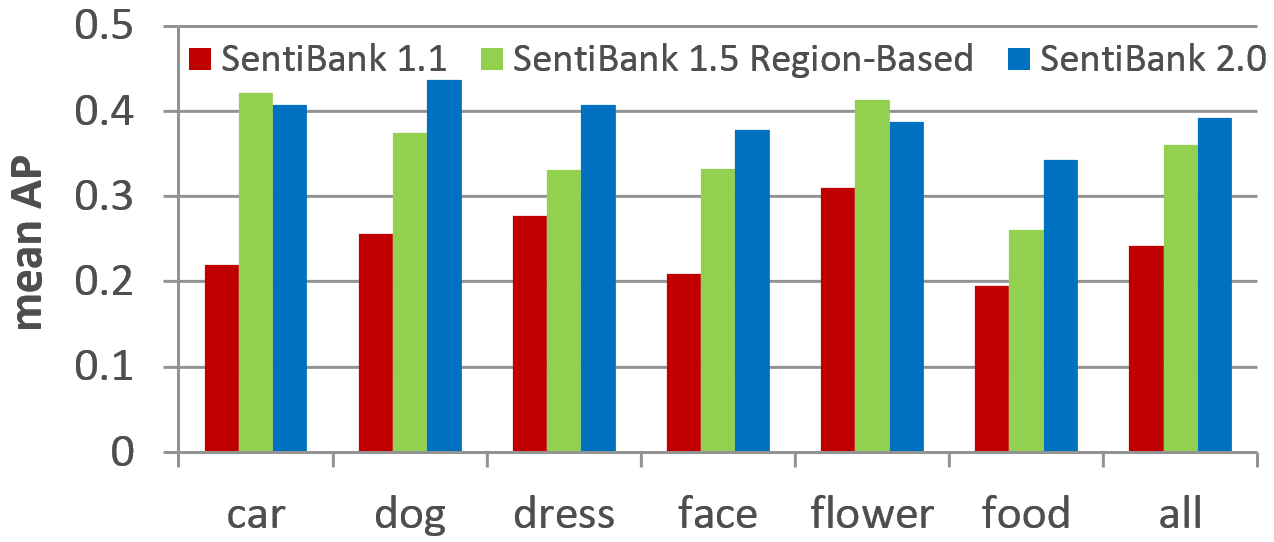}\\
  \caption{The mean AP for each and all noun categories for the subset of 135 ANPs mentioned in \secref{dataset}. }\label{fig:ap}
\end{figure}

The retrieval performance is evaluated on the subset of 135 ANPs mentioned in \secref{dataset}. We apply the models trained from SentiBank 1.1, 1.5R and DeepSentiBank to the test set. For each ANP, the test images are ranked by the estimated probability of the ANP. The performance is measured by average precision (AP) at top 20. The mean AP for each and all noun categories are shown in \figref{ap}. Although not designed for retrieval, DeepSentiBank still outperforms SentiBank 1.1 by 62.3\% and SentiBank 1.5R by 8.9\%. Note DeepSentiBank is only trained on whole images and does not consider concept localization or concept similarity. It means the performance could be further improved if we incorporate the two factors into deep learning. Recently, R-CNN \cite{girshick2014rich} shows state-of-the-art performance on object detection, which can be a promising candidate approach for the concept localization.

\section{Conclusion}

This paper presents a visual sentiment concept classification model based on deep convolutional neural networks. The deep CNNs model is trained based on Caffe, a newly developed deep learning framework. To deal with the biased training data which only contains images with strong sentiment and to prevent overfitting, we initialize the model with the model weights trained from ImageNet. Performance evaluation shows the newly trained deep CNNs model DeepSentiBank is significantly better in both annotation and retrieval, compared to previous work using independent binary SVM classification models.
In the future, we will incorporate the concept localization into the deep CNNs model, and improve network structure by leveraging concept relations.
The high performance boost will also help to improve applications built on SentiBank, such as assistive comment robot \cite{Chen2014Predicting} and twitter sentiment prediction, or other applications such as  sentiment-aware image editing.

\begin{table*}
  \centering
  \caption{Evaluation of Different SentiBank Models in terms of Retrieval}\label{tab:accuracy}
  \vspace{0.1in}

  \begin{tabular}{|c|c|c|c|c|c|c|}
\hline
\multirow{2}{*}{SentiBank ver.} &  \multicolumn{3}{|c|}{2,089 ANPs} & \multicolumn{3}{|c|}{1,200 ANPs}
\\
\cline{2-7}
&  Top-1 & Top-5 & Top-10 &  Top-1 & Top-5 & Top-10
\\
\hline
\hline
 SentiBank 1.1 & 1.7075\% &	6.3211\%	& 10.2917\% & 3.0386\% &	11.4288\% &	 18.7356\%
\\
\hline
 DeepSentiBank w/o fine-tuning  & 6.5235\%	&	16.0095\%	&	22.4941\%	&	 11.4430\%	&	28.4856\%	&	39.0800\%	
\\
\hline
 DeepSentiBank & \textbf{8.1629\%}	&	\textbf{19.1132\%}	&	 \textbf{26.1012\%}	& \textbf{14.3572\%}	&	\textbf{33.3726\%}	&	 \textbf{44.3664\%}	

\\
\hline
\end{tabular}
\end{table*}

\bibliographystyle{plain}
\bibliography{multimedia}
\end{document}